\documentclass[runningheads]{llncs}
\usepackage{graphicx}
\usepackage[utf8]{inputenc} 
\usepackage[T1]{fontenc}    
\usepackage{hyperref}       
\usepackage{url}            
\usepackage{booktabs}       
\usepackage{amsfonts}       
\usepackage{nicefrac}       
\usepackage{microtype}      
\usepackage{longtable}

\usepackage{amsmath,amssymb}

\newcommand{\mE}{\mathbb{E}}
\newcommand{\mV}{\mathbb{V}}

\newcommand{\Eqref}[1]{Eq. (\ref{#1})}

\usepackage{algorithm}
\usepackage{algpseudocode}

\algnewcommand{\Set}[1]{%
  \textbf{Set:}\\
}

\usepackage{graphicx}
\usepackage[caption=false]{subfig}

\usepackage{tabularx}

\usepackage{color}
\definecolor{dkgreen}{rgb}{0,0.6,0}
\definecolor{customgray}{rgb}{0.25,0.25,0.25}
\definecolor{customred}{rgb}{0.8,0.05,0.05}
\definecolor{customblue}{rgb}{0.05,0.05,0.8}

\usepackage{color}

\begin{document}
%
\title{Towards a Principled Learning Rate Adaptation for Natural Evolution Strategies}

\titlerunning{Towards a Principled Learning Rate Adaptation for NES}
%
\author{Masahiro Nomura\inst{1} \and
Isao Ono\inst{2}}
\authorrunning{Nomura and Ono.}
%

\institute{Tokyo Institute of Technology, Japan
\email{nomura.m.ad@m.titech.ac.jp}
\and
Tokyo Institute of Technology, Japan
\email{isao@c.titech.ac.jp}}
\maketitle              
\begin{abstract}
Natural Evolution Strategies (NES) is a promising framework for black-box continuous optimization problems.
NES optimizes the parameters of a probability distribution based on the estimated natural gradient, and one of the key parameters affecting the performance is the learning rate.
We argue that from the viewpoint of the natural gradient method, \emph{the learning rate should be determined according to the estimation accuracy of the natural gradient}.
To do so, we propose a new learning rate adaptation mechanism for NES.
The proposed mechanism makes it possible to set a high learning rate for problems that are relatively easy to optimize, which results in speeding up the search.
On the other hand, in problems that are difficult to optimize (e.g., multimodal functions), the proposed mechanism makes it possible to set a conservative learning rate when the estimation accuracy of the natural gradient seems to be low, which results in the robust and stable search.
The experimental evaluations on unimodal and multimodal functions demonstrate that the proposed mechanism works properly depending on a search situation and is effective over the existing method, i.e., using the fixed learning rate.

\end{abstract}

\section{Introduction}
\label{sec:intro}

Natural Evolution Strategies (NES)~\cite{wierstra2014natural,glasmachers2010exponential,sun2009efficient,yi2009stochastic} is a promising framework for black-box continuous optimization problems.
Instead of directly seeking the optimal solution $x^{*}$, NES optimizes the parameter $\theta$ of a probability distribution $p(x | \theta)$.
The expectation of the objective function $f(x)$ over the solution space $J(\theta) = \int f(x) p(x | \theta) dx$ is minimized by repeatedly updating the parameter of the probability distribution based on the estimated natural gradient~\cite{amari1998why}.
In this study, we focus on NES using a multivariate normal distribution as the probability distribution.
The natural gradient plays an important role in evolution strategies and randomized algorithms:
For example, the rank-$\mu$ update~\cite{hansen2003reducing,hansen2004evaluating} of CMA-ES~\cite{hansen2006cma} can also be regarded as using the estimated value of the natural gradient~\cite{akimoto2010bidirectional}.
Information Geometric Optimization \cite{ollivier2017information}, which is a generalized framework of NES and the rank-$\mu$ update of CMA-ES, has been actively studied in recent years~\cite{akimoto2014comparison,beyer2014convergence,otwinowski2020information}.

As with other evolution strategies, one of the critical parameters in NES is a learning rate for the parameter of the probability distribution.
If the learning rate is too high, the parameter update will be unstable and the performance will deteriorate.
On the other hand, if the learning rate is too low, the speed of approaching the optimal solution will be slow, resulting in poor performance.
Therefore, setting an appropriate learning rate is essential for maximizing the performance of NES.

There are a few studies on learning rate adaptation of NES.
DX-NES proposed by Fukushima et al. switches the learning rate based on the norm of the evolution path which accumulates the movement of the \emph{normalized} mean vector~\cite{fukushima2011proposal}.
The effectiveness of switching the learning rate in DX-NES is demonstrated empirically.
In fact, the recently proposed DX-NES variants~\cite{nomura2021distance,nomura2021natural}, which also employ switching of the learning rate, show promising performance on unconstrained and implicitly constrained black-box optimization problems.
Another learning rate adaptation method based on maximum likelihood estimation is proposed in the literature of the CMA-ES~\cite{loshchilov2014maximum}.

In this paper, we propose a new learning rate adaptation mechanism in view of the natural gradient method;
our work is based on a principle that \emph{the learning rate of the natural gradient method should depend on its estimation accuracy}.
To measure the estimation accuracy of the natural gradient, we calculate the movement of Kullback-Leibler (KL) divergence, which was first introduced in the population size adaptation of the CMA-ES~\cite{nishida2016population,nishida2018psa}.
We extend the notion to the learning rate adaptation in NES.

The aim of this study is to understand the behavior of NES with the learning rate adaptation mechanism, rather than developing a method that achieves state-of-the-art performance, as this is the first work of learning rate adaptation based on the estimation accuracy of the natural gradient in NES.
To that end, we decide to incorporate the learning rate adaptation into xNES~\cite{glasmachers2010exponential}, which is a simple and promising variant of NES.

The rest of this paper is organized as follows.
In Section~\ref{sec:xnes}, we describe xNES algorithm.
In Section~\ref{sec:lra}, we propose a learning rate adaptation mechanism based on the estimation accuracy of the natural gradient.
In Section~\ref{sec:exp}, we experiment on unimodal and multimodal benchmark problems to investigate the effect of the learning rate adaptation mechanism. 
Section~\ref{sec:con} concludes with summary and future direction of this work.

\section{xNES}
\label{sec:xnes}
xNES~\cite{glasmachers2010exponential} uses a multivariate normal distribution $\mathcal{N}(m, \sigma^2 B B^{\top})$ as the probability distribution.
Here, $m \in \mathbb{R}^d$ is the mean vector, $\sigma \in \mathbb{R}_{>0}$ is the step-size, and $B \in \mathbb{R}^{d\times d}$ is the normalization transformation matrix where $\det(B) = 1$.
The update of xNES is performed by using an estimated natural gradient in the parameter space of the multivariate normal distribution.

xNES first initializes the parameters $m, \sigma$, and $B$.
Then, the following steps are repeated until a stopping criterion is met.

\noindent
{\bf Step 1. Sampling and Sorting}\\
For $i \in \{1, \cdots, \lambda \}$, sample $\lambda$ solutions $x_i \sim \mathcal{N}(m, \sigma^2 B B^{\top})$ as follows.
Generate $d$-dimensional standard normal vectors $z_i \sim \mathcal{N} (0, I)$
and compute $x_i = m + \sigma B z_i$.
Evaluate the generated solutions on the objective function and obtain their objective values.
Then, sort the solutions according to their evaluation values.

\noindent
{\bf Step 2. Estimating Natural Gradient}\\
Estimate the natural gradient based on the sorted solutions as follows:
\begin{align*}
  G_{\delta} &= \sum_{i=1}^{\lambda} w_i z_i, G_M = \sum_{i=1}^{\lambda} w_i(z_i z_i^{\top} - I),\\
  G_{\sigma} &= {\rm Tr}(G_M) / d, G_{B} = G_M - G_{\sigma} \cdot I,
\end{align*}
where
$w_i$ is the weight function
\begin{align*}
w_i = \frac{\max \left(0, \ln\left(\frac{\lambda}{2} + 1 \right) - \ln{(i)} \right)}{\sum_{j=1}^\lambda \max \left(0, \ln\left(\frac{\lambda}{2} + 1 \right) - \ln{(i)} \right)} - \frac{1}{\lambda}.
\end{align*}
The weight function holds $\sum_{i=1}^{\lambda} w_i = 0$.
Note that xNES uses the weight function instead of using raw evaluation values.
This technique is called \emph{fitness shaping}, and it improves the robustness of the algorithm due to the invariance for the monotone transformation of the objective function and enables linear convergence~\cite{beyer2014convergence}.

\noindent
{\bf Step 3. Updating Parameters}\\
Based on the estimated natural gradient, update the parameters of the multivariate normal distributions as follows:
\begin{align*}
  m &\gets m + \eta_{m} \sigma B G_{\delta},\\
  \sigma &\gets \sigma \cdot \exp (\eta_{\sigma} / 2 \cdot G_{\sigma}),\\
  B &\gets B \cdot \exp(\eta_B / 2 \cdot G_B),
\end{align*}
where $\eta_{m}, \eta_{\sigma}$, and $\eta_B$ are the learning rates for updating $m$, $\sigma$, and $B$, respectively.
These learning rates above have default values~\cite{glasmachers2010exponential};
$\eta_m = 1$ and $\eta_{\sigma} = \eta_B = \frac{3}{5} \cdot \frac{(3 + \log(d))}{d\sqrt{d}}$.

\section{Learning Rate Adaptation}
\label{sec:lra}

\subsection{Motivation}
While default values of the learning rates are presented in xNES, Fukushima et al. have pointed out that the default values are too conservative in a certain situation and there is much room for improvement~\cite{fukushima2011proposal}.
However, simply increasing the learning rate causes performance degradation in problems where it is difficult to estimate the natural gradient.
It is thus important to adapt the learning rate according to the search situation in order to maximize the performance of xNES.

In this work, we try to adapt the learning rates $\eta_{\sigma}$ and $\eta_B$.
That is, we focus on only the learning rates related to the covariance matrix.
We fix $\eta_m = 1$, which is the default value presented in~\cite{glasmachers2010exponential} and widely used in the literature of the CMA-ES~\cite{hansen2003reducing,hansen2016cma} as well.

To this end, we introduce a learning rate adaptation mechanism that dynamically adapts the learning rates based on the estimation accuracy of the natural gradient.
To quantify the estimation accuracy of the natural gradient, we introduce an evolution path in the \emph{parameter space}~\cite{nishida2018psa}, which accumulates successive parameter movements.
The length of the evolution path in the parameter space, described in detail in Section~\ref{sec:evo_path_cov}, is used to measure the accuracy of the natural gradient.
We believe that, if the length of the evolution path is larger than its expectation under a random function, the accuracy is high, as the tendency of parameter update can be captured.
On the contrary, we believe that, if the length of the evolution path is close to its expectation under a random function, the estimation is dominated by noise, and the accuracy is low.

In this study, we consider an evolution path in the parameter space of only the covariance matrix, not the mean vector, because the learning rate for the mean vector is fixed.
This is different from existing studies that use the evolution path in the parameter space~\cite{nishida2016population,nishida2018psa}.
We will investigate the behavior of the evolution path in Section~\ref{sec:exp}.

\subsection{Evolution Path for Covariance Matrix}
\label{sec:evo_path_cov}
In this work, we introduce an evolution path in the parameter space of the covariance matrix to quantify the estimation accuracy of the natural gradient.
We use a modification of the evolution path proposed in~\cite{nishida2018psa}, which considers both the mean vector and the covariance matrix.
Let $\Sigma := \sigma^2 B B^{\top}$.
The covariance movement matrix $\delta \Sigma^{(t+1)}$ is defined to capture the movement of the covariance matrix from iteration $t$ to $t+1$, which is updated as
\begin{align}
    \delta \Sigma^{(t+1)} &= (\sigma^{(t+1)})^2 B^{(t+1)} {B^{(t+1)}}^{\top} - (\sigma^{(t)})^2 B^{(t)} {B^{(t)}}^{\top}.
\end{align}

We then define the evolution path in the parameter space of the covariance matrix.
\begin{align}
    \label{eq:evo_path_theta}
    p_{\Sigma}^{(t+1)} &= (1-\beta) p_{\Sigma}^{(t)} + \sqrt{\beta(2-\beta)} \frac{\mathcal{I}_{\Sigma^{(t)}}^{\frac{1}{2}} \delta \Sigma^{(t+1)}}{\mE[\| \mathcal{I}_{\Sigma^{(t)}}^{\frac{1}{2}} \delta \Sigma^{(t+1)} \|^2]^{\frac{1}{2}}},
\end{align}
where $\beta$ is a cumulation factor of the evolution path and $\mathcal{I}_{\Sigma^{(t)}}$ is the Fisher information matrix of the covariance matrix of the multivariate normal distribution.
The expectation $\mE[\cdot]$ is taken under a random function $f(x) = \epsilon$, where $\epsilon$ is independently drawn from the identical distribution for each evaluation.
We use the approximation of $\mE[\| \mathcal{I}_{\Sigma^{(t)}}^{\frac{1}{2}} \delta \Sigma^{(t+1)} \|^2]^{\frac{1}{2}}$, which will be derived in Section~\ref{sec:approx_kl_Sigma}.

Using the result from Eq.~(21) and Appendix B in~\cite{nishida2016population}, we define the \emph{length} of the evolution path $p_{\Sigma}$, which represents the movement of the KL divergence in the parameter space of the covariance matrix, as follows:
\begin{align}
    l_{\theta}^{(t+1)} := \frac{{\rm Tr}\left(\left(p_{\Sigma}^{(t+1)}\right)^2 \right)}{2}.
\end{align}
Although we do not consider the movement of the KL divergence in the parameter space of the mean vector, we use $\theta$ as a notation for the parameter space of the probability distribution.

Under a random function, the length of the evolution path approaches $1$ as the iteration $t$ increases.
Therefore, comparing the the length of the evolution path with \emph{the normalization factor} $\gamma_{\theta}^{(t+1)}$ which is updated as
\begin{align}
    \gamma_{\theta}^{(t+1)} &= (1-\beta)^2 \gamma_{\theta}^{(t)} + \beta(2-\beta),
\end{align}
we can obtain the estimation of the accuracy of the parameter update.
The initial parameter $\gamma_{\theta}^{(0)}$ is set to $\gamma_{\theta}^{(0)} = 0$.

\subsection{Updating Learning Rate}
\label{sec:update_lr}
In this section, we give a procedure for the learning rate adaptation.
As described, we argue that the learning rate should depend on the estimation accuracy of the natural gradient.
When the accuracy is high, the learning rate should be increased, and when the accuracy is low, the learning rate should be decreased.

The learning rate adaptation is performed as follows:
\begin{align}
    \eta_{\sigma}^{(t+1)} &= \eta_{\sigma}^{(t)} \exp \left( \beta_{\sigma} \left( \frac{l_{\theta}^{(t+1)}}{\alpha_{\sigma}} - \gamma_{\theta}^{(t+1)} \right) \right),\\
    \eta_B^{(t+1)} &= \eta_B^{(t)} \exp \left( \beta_B \left(  \frac{l_{\theta}^{(t+1)}}{\alpha_B} - \gamma_{\theta}^{(t+1)} \right) \right),
\end{align}
where $\alpha_{\sigma}, \alpha_B, \beta_{\sigma}$, and $\beta_B$ are pre-defined hyperparameters.
It is possible to set different hyperparameters for $\eta_{\sigma}$ and $\eta_{B}$, respectively, if needed.
In this study, we employ the same value for easier interpretation, i.e., $\alpha_{\sigma} = \alpha_B$ and $\beta_{\sigma} = \beta_B$.

We clip the learning rates to prevent them from being updated to unexpected ranges by the following equations:
\begin{align}
    \eta_{\sigma}^{(t+1)} &\gets \mathrm{clip} (\eta_{\sigma}^{(t+1)}, \eta_{\sigma}^{\rm min}, \eta_{\sigma}^{\rm max}), \\
    \eta_B^{(t+1)} &\gets \mathrm{clip} (\eta_B^{(t+1)}, \eta_B^{\rm min}, \eta_B^{\rm max}),
\end{align}
where $\eta_{\sigma}^{\rm max}$ and $\eta_{\sigma}^{\rm min}$ are the maximum and the minimum values of the learning rate for step-size $\sigma$, respectively.
Similarly, $\eta_{B}^{\rm max}$ and $\eta_{B}^{\rm min}$ are the maximum and the minimum values of the learning rate for the normalized transformation matrix $B$, respectively.
The $\mathrm{clip}$ function is defined as $\mathrm{clip}(u, a, b) := \min ( \max (u, a), b)$.

To prevent extrapolation in the update of the parameter, we set the maximum value of the learning rates to $1$, i.e., $\eta_{\sigma}^{\rm max} = \eta_{B}^{\rm max} = 1$.
Also, the minimum value of the learning rates is set to the default value of xNES, as it is pointed out that the setting of the learning rates in xNES is often too conservative~\cite{fukushima2011proposal}.
Therefore, we use the values recommended in~\cite{glasmachers2010exponential} for $\eta_{\sigma}^{\rm min}$ and $\eta_{B}^{\rm min}$,
i.e., $\eta_{\sigma}^{\rm min} = \eta_{B}^{\rm min} = \frac{3}{5} \cdot \frac{(3 + \log(d))}{d\sqrt{d}}$.

\subsection{Approximation of $\mE[\| \mathcal{I}_{\Sigma^{(t)}}^{\frac{1}{2}} \delta \Sigma^{(t+1)} \|^2]^{\frac{1}{2}}$}
\label{sec:approx_kl_Sigma}

In this section, we derive an approximation of $\mE[\| \mathcal{I}_{\Sigma^{(t)}}^{\frac{1}{2}} \delta \Sigma^{(t+1)} \|^2]^{\frac{1}{2}}$, which represents a change of the KL divergence in terms of the covariance matrix.

Let $C^{(t)} = B^{(t)} {B^{(t)}}^{\top}, \delta \Sigma = {\sigma^{(t+1)}}^2 C^{(t+1)} - {\sigma^{(t)}}^2 C^{(t)}$, $\Sigma^{-1} = {\sigma^{(t)}}^{-2} {C^{(t)}}^{-1}$, $\delta \sigma = \sigma^{(t+1)} / \sigma^{(t)}$, and $\delta C = C^{(t+1)} - C^{(t)}$.
To derive the approximation, we use the Slepian-Bangs formula~\cite{slepian1954estimation,bangs1971array} and obtain $\| \mathcal{I}_{\Sigma^{(t)}}^{\frac{1}{2}} \delta \Sigma \|^2 = \delta \Sigma^{\top} \mathcal{I}_{\Sigma^{(t)}} \delta \Sigma = \frac{1}{2} {\rm Tr}\left( \delta \Sigma \Sigma^{-1} \delta \Sigma \Sigma^{-1} \right)$.
We will derive the expectation of this equation.
From the result provided by Nishida and Akimoto~\cite{nishida2018psa}, we can obtain $\mE[ {\rm Tr}\left( \delta \Sigma \Sigma^{-1} \delta \Sigma \Sigma^{-1} \right)] = \mE[\delta \sigma^4] {\rm Tr} \left( \mE \left[ \left( {C^{(t)}}^{-1/2} \cdot \delta C \cdot {C^{(t)}}^{-1/2} \right)^2 \right] \right) + d (\mE [\delta \sigma^4] - 2 \mE[\delta \sigma^2])$.
We then need to derive the approximation of $\mE[\delta \sigma^4], \mE[\delta \sigma^2]$, and ${\rm Tr} \left( \mE \left[ \left( {C^{(t)}}^{-1/2} \cdot \delta C \cdot {C^{(t)}}^{-1/2} \right)^2 \right] \right)$.

\noindent
\textbf{Derivation of $\mE[ \delta \sigma^a ] \ (a = 2, 4)$:}

The update equation of step-size in xNES can be rewritten as
{\small
\begin{align*}
    \sigma^{(t+1)} = \sigma^{(t)} \cdot \exp \left( \frac{\eta_{\sigma}}{2d} \left( \sum_{j=1}^d \sum_{i=1}^{\lambda} w_i \left( [z_i]_j^2 - 1 \right) \right) \right).
\end{align*}
}
Then, by the second order Taylor expansion, for any $a \in \mathbb{R}$,
{\small
\begin{align*}
    \mE[ \delta \sigma^a ] &= \mE \left[ \exp \left( a \frac{\eta_{\sigma}}{2d} \left( \sum_{j=1}^d \sum_{i=1}^{\lambda} w_i \left( [z_i]_j^2 - 1 \right) \right) \right) \right] \\
    &\approx 1 + a \cdot \frac{\eta_{\sigma} \lambda}{2} \mE [w_i ([z_i]_j^2 - 1)] + \frac{a^2}{2} \cdot \left( \frac{\eta_{\sigma}}{2d} \right)^2 \mE \left[ \left( \sum_{j=1}^d \sum_{i=1}^{\lambda} w_i \left( [z_i]_j^2 - 1 \right) \right)^2 \right].
\end{align*}
}
We will thus calculate the expectations in the above equation.
First, from $\mE[[z_i]^2] = \mV[[z_i]] + \mE[[z_i]] = 1$, $\mE [w_i ([z_i]_j^2 - 1)] = 0$.
Next, noting that $\mV \left[ z^2 \right] = 2$ and the independence,
{\small
\begin{align*}
    \mE \left[ \left( \sum_{j=1}^d \sum_{i=1}^{\lambda} w_i \left( [z_i]_j^2 - 1 \right) \right)^2 \right] &= \left( \underbrace{\mE \left[\sum_{j=1}^d \sum_{i=1}^{\lambda} w_i \left( [z_i]_j^2 - 1 \right) \right]}_{=0} \right)^2 + \mV \left[ \sum_{j=1}^d \sum_{i=1}^{\lambda} w_i \left( [z_i]_j^2 - 1 \right) \right] \\
    &= \sum_{j=1}^d \sum_{i=1}^{\lambda} w_i^2 \mV[[z_i]^2_j] = 2d / \mu_{w},
\end{align*}
}
where $\mu_{w} = \sum_{i=1}^{\lambda} 1 / w_i^2$.
By combining these results, $\mE[\delta \sigma^a] \approx 1 + \frac{a^2 \eta_{\sigma}^2}{4 d \mu_{w}}$.
We thus obtain
{\small
\begin{align*}
    &\mE[\delta \sigma^4] \approx 1 + \frac{4 \eta_{\sigma}^2}{d \mu_w}, \mE[\delta \sigma^2] \approx 1 + \frac{\eta_{\sigma}^2}{d \mu_w},\\
    &\mE[\delta \sigma^4] - 2 \mE[\delta \sigma^2] \approx \frac{2 \eta_{\sigma}^2}{d \mu_w} - 1.
\end{align*}
}

\noindent
\textbf{Derivation of ${\rm Tr} \left( \mE \left[ \left( {C^{(t)}}^{-1/2} \cdot \delta C \cdot {C^{(t)}}^{-1/2} \right)^2 \right] \right)$:}

Let $\Delta = {\rm Tr} \left( \sum_{i=1}^{\lambda} w_i (z_i z_i^{\top} - I) \right)$.
The first order Taylor expansion of $C$ in xNES can be obtained as
{\small
\begin{align*}
    C^{(t+1)} \approx C^{(t)} + \eta_B {C^{(t)}}^{1/2} \left( \Delta - \frac{{\rm Tr}(\Delta)}{d}I \right) {C^{(t)}}^{1/2}.
\end{align*}
}
From $\delta C \approx \eta_B {C^{(t)}}^{1/2} \left( \Delta - \frac{{\rm Tr}(\Delta)}{d}I \right) {C^{(t)}}^{1/2}$, ${C^{(t)}}^{-1/2} \delta C {C^{(t)}}^{-1/2} \approx \eta_B \left( \Delta - \frac{{\rm Tr}(\Delta)}{d}I \right)$,
{\small
\begin{align*}
    &{\rm Tr} \left( \mE \left[ \left( {C^{(t)}}^{-1/2} \cdot \delta C \cdot {C^{(t)}}^{-1/2} \right)^2 \right] \right) \\
    &\approx {\rm Tr} \left( \mE \left[ \left( \eta_B \left( \Delta - \frac{{\rm Tr}(\Delta)}{d}I \right) \right)^2 \right] \right) \\
    &= \mE \left[ {\rm Tr} \left( \eta_B \left( \Delta - \frac{{\rm Tr}(\Delta)}{d}I \right) \right)^2 \right] = \eta_B^2 \left\{ \mE \left[ {\rm Tr}(\Delta^2) \right] - \frac{1}{d} \mE [{\rm Tr} (\Delta)^2] \right\}.
\end{align*}
}

\noindent
\textbf{Derivation of $\mE \left[ {\rm Tr}(\Delta^2) \right]$ and $\mE [{\rm Tr} (\Delta)^2]$:}

{\small
\begin{align*}
    &\mE \left[ {\rm Tr}(\Delta^2) \right] = \mE \left[ {\rm Tr} \left( \sum_{i=1}^{\lambda} \sum_{j=1}^{\lambda} w_i w_j (z_i z_i^{\top} - I) (z_j z_j^{\top} - I) \right) \right] \\
    &= \sum_{i=1}^{\lambda} \sum_{j=1}^{\lambda} w_i w_j \mE \left[ {\rm Tr} \left( (z_i z_i^{\top} - I) (z_j z_j^{\top} - I) \right) \right] \\
    &= \sum_{i=1}^{\lambda} \sum_{j=1}^{\lambda} w_i w_j \mE \left[ {\rm Tr} \left( z_i z_i^{\top} z_j z_j^{\top} \right) - {\rm Tr}(z_i z_i^{\top}) - {\rm Tr}(z_j z_j^{\top}) + {\rm Tr}(I) \right] \\
    &= \sum_{i=1}^{\lambda} \sum_{j=1}^{\lambda} w_i w_j (\mE [(z_i^{\top} z_j)^2] - \mE [z_i^{\top} z_i] - \mE [z_j^{\top} z_j] + d).
\end{align*}
}

Note that $\mE [z_i^{\top} z_i] = d$.
Then, $\forall i,j \geq 1 (i \neq j)$,
$\mE [(z_i^{\top} z_j)^2] = d, \mE [(z_i^{\top} z_i)^2] = d^2 + 2d.$
Therefore,
{\small
\begin{align*}
    \mE \left[ {\rm Tr}(\Delta^2) \right] &= \sum_{i=1}^{\lambda} w_i^2 (d^2 + 2d -d -d + d) + \sum_{i,j: i \neq j} w_i w_j (\underbrace{d - d - d + d}_{=0})\\
    &= \sum_{i=1}^{\lambda} w_i^2 (d^2 + d) = (d^2 + d) / \mu_{w}.
\end{align*}
}

We then derive $\mE \left[{\rm Tr} (\Delta)^2 \right]$.
First, $\Delta = {\rm Tr} \left( \sum_{i=1}^{\lambda} w_i (z_i z_i^{\top} - I) \right)$ is written as
{\small
\begin{align*}
    {\rm Tr} \left( \sum_{i=1}^{\lambda} w_i (z_i z_i^{\top} - I) \right) &= \sum_{i=1}^{\lambda} w_i ({\rm Tr} (z_i z_i^{\top}) - d) \\
    &= \sum_{i=1}^{\lambda} w_i {\rm Tr} (z_i z_i^{\top}) \\
    &= \sum_{i=1}^{\lambda} w_i \| z_i \|^2.
\end{align*}
}
In the second line, we used $\sum_{i=1}^{\lambda} w_i = 0$. 
Then, $\mE [{\rm Tr} (\Delta)^2] = \mE [ \sum_{i=1}^{\lambda} w_i^2 \| z_i \|^4 + \sum_{i,j: i \neq j} w_i w_j \mE [\| z_i \|^2] \mE [\| z_j \|^2] = (d^2 + 2d) / \mu_w - d^2 / \mu_w = 2d / \mu_w$.
We used $\sum_{i,j: i \neq j} w_i w_j = \sum_{i,j} w_i w_j - \sum_{i=1}^{\lambda} w_i^2 = - 1 / \mu_w$ due to $\sum_{i=1}^{\lambda} w_i = 0$.

By combining these results,
{\small
\begin{align*}
    {\rm Tr} \left( \mE \left[ \left( {C^{(t)}}^{-1/2} \cdot \delta C \cdot {C^{(t)}}^{-1/2} \right)^2 \right] \right) = \frac{\eta_B^2}{\mu_w} (d^2 + d - 2).
\end{align*}
}

\noindent
\textbf{Approximation Result:}

From the results above,
{\small
\begin{align*}
    &\mE[ {\rm Tr}\left( \delta \Sigma \Sigma^{-1} \delta \Sigma \Sigma^{-1} \right)] \\
    &= \underbrace{\mE [\delta \sigma^4]}_{\approx 1 + \frac{4 \eta_{\sigma}^2}{d \mu_w}} \underbrace{{\rm Tr} \left( \mE \left[ \left( {C^{(t)}}^{-1/2} \cdot \delta C \cdot {C^{(t)}}^{-1/2} \right)^2 \right] \right)}_{\approx \eta_B^2(d^2 + d - 2) / \mu_w} + d (\underbrace{\mE [\delta \sigma^4] - 2 \mE [\delta \sigma^2]}_{\approx \frac{2 \eta_{\sigma}^2}{d \mu_w} - 1}) + d \\
    &\approx \frac{1}{\mu_w} \left\{ \eta_B^2 \left( 1 + \frac{4 \eta_{\sigma}^2}{d \mu_w} \right) (d^2 + d - 2) + 2 \eta_{\sigma}^2 \right\}.
\end{align*}
}

Therefore,
\begin{align}
    \label{eq:approx_final_kl_diff}
    &\mE[\| \mathcal{I}_{\Sigma^{(t)}}^{\frac{1}{2}} \delta \Sigma^{(t+1)} \|^2] \approx \frac{1}{\mu_w} \left\{ \frac{\eta_B^2}{2} \left( 1 + \frac{4 \eta_{\sigma}^2}{d \mu_w} \right) (d^2 + d - 2) + \eta_{\sigma}^2 \right\}.
\end{align}
We recalculate this approximation every iteration because it depends on the dynamically changing learning rates, $\eta_{\sigma}$ and $\eta_B$.

\subsection{Overall Procedure}
The overall procedure of xNES with the proposed learning rate adaptation mechanism is shown in Algrithm~\ref{alg:xnes_lra}.
The parameters $\eta_{\sigma}^{\rm def}$ and $\eta_{B}^{\rm def}$ are the recommended setting of the learning rates in~\cite{glasmachers2010exponential}, and $O$ is the zero matrix.
The procedures in line 3-14 are the same as xNES.
In line 15, the covariance movement matrix is updated.
In line 16, the expectation of the length of the evolution path under a random function is approximated by using Eq.~(\ref{eq:approx_final_kl_diff}).
In line 17, the evolution path in the parameter space of the covariance matrix is updated.
In line 18, the length of the evolution path is calculated.
In line 19, the normalization factor for the evolution path is updated.
In line 20-21, the learning rates for the step-size and the normalization transformation matrix are updated with clipping.

\begin{algorithm}
\caption{xNES with the learning rate adaptation.}
\label{alg:xnes_lra}
\begin{algorithmic}[1]
\Require $m^{(0)} \in \mathbb{R}^d, \sigma^{(0)} \in \mathbb{R}_{+}, B^{(0)} \mathbb{R}^{d\times d}, \lambda \in \mathbb{N}$
\Require $\alpha_{\sigma}, \alpha_{B}, \beta_{\sigma}, \beta_{B}, \eta_{\sigma}^{\rm min}, \eta_B^{\rm min}, \eta_{\sigma}^{\rm max}, \eta_{B}^{\rm max}$
\State $t = 0, p_{\Sigma}^{(0)} = O, \gamma_{\theta}^{(0)} = 0, \eta_{\sigma}^{(0)} = \eta_{\sigma}^{\rm def}, \eta_{B}^{(0)} = \eta_{B}^{\rm def}, \eta_m = 1$
\While{stopping criterion not met}
  \For{$i \in \{ 1, \cdots, \lambda \}$}
    \State $z_i \sim \mathcal{N}(0, I)$
    \State $x_i = m^{(t)} + \sigma^{(t)} B^{(t)} z_i$
  \EndFor
  \State Evaluate the solutions and sort $\{ (z_i, x_i)\}$
  
  \State $G_{\delta} = \sum_{i=1}^{\lambda} w_i z_i$
  \State $G_M = \sum_{i=1}^{\lambda} w_i(z_i z_i^{\top} - I)$
  \State $G_{\sigma} = {\rm Tr}(G_M) / d$
  \State $G_{B} = G_M - G_{\sigma} \cdot I$
  
  \State $m^{(t+1)} = m^{(t)} + \eta_{m} \sigma^{(t)} B^{(t)} G_{\delta}$
  \State $\sigma^{(t+1)} = \sigma^{(t)} \cdot \exp (\eta_{\sigma}^{(t)} / 2 \cdot G_{\sigma})$
  \State $B^{(t+1)} = B^{(t)} \cdot \exp(\eta_B^{(t)} / 2 \cdot G_B)$
  
  \State $\delta \Sigma^{(t+1)} = (\sigma^{(t+1)})^2 B^{(t+1)} {B^{(t+1)}}^{\top} - (\sigma^{(t)})^2 B^{(t)} {B^{(t)}}^{\top}$
  \State Approximate $\mE[\| \mathcal{I}_{\Sigma^{(t)}}^{\frac{1}{2}} \delta \Sigma^{(t+1)} \|^2]^{\frac{1}{2}}$ by Eq.~(\ref{eq:approx_final_kl_diff})
  \State $p_{\Sigma}^{(t+1)} = (1-\beta) p_{\Sigma}^{(t)} + \sqrt{\beta(2-\beta)} \frac{\mathcal{I}_{\Sigma^{(t)}}^{\frac{1}{2}} \delta \Sigma^{(t+1)}}{\mE[\| \mathcal{I}_{\Sigma^{(t)}}^{\frac{1}{2}} \delta \Sigma^{(t+1)} \|^2]^{\frac{1}{2}}}$
  \State $l_{\theta}^{(t+1)} = {\rm Tr}({p_{\Sigma}^{(t+1)}}^2) / 2$
  \State $\gamma_{\theta}^{(t+1)} = (1-\beta)^2 \gamma_{\theta}^{(t)} + \beta(2-\beta)$
  
  \State $\eta_{\sigma}^{(t+1)} = \mathrm{clip} \left( \eta_{\sigma}^{(t)} \exp \left( \beta_{\sigma} \left( \frac{ l_{\theta}^{(t+1)} }{\alpha_{\sigma}} - \gamma_{\theta}^{(t+1)} \right) \right), \eta_{\sigma}^{\rm min}, \eta_{\sigma}^{\rm max} \right)$
  \State $\eta_B^{(t+1)} = \mathrm{clip} \left( \eta_B^{(t)} \exp \left( \beta_B \left(  \frac{ l_{\theta}^{(t+1)} }{\alpha_B} - \gamma_{\theta}^{(t+1)} \right) \right), \eta_B^{\rm min}, \eta_B^{\rm max} \right)$

  \State $t \leftarrow t + 1$
\EndWhile

\end{algorithmic}
\end{algorithm}

\section{Experiments}
\label{sec:exp}
In this section, we investigate the following research questions (RQs).

\begin{itemize}
    \item [\textbf{RQ1.}] When the learning rate is fixed, how does the evolution path in \Eqref{eq:evo_path_theta} of xNES behave on unimodal and multimodal functions?
    \item [\textbf{RQ2.}] How is the learning rate adapted in xNES with the proposed learning rate adaptation mechanism?
    \item [\textbf{RQ3.}] Does xNES with the proposed learning rate adaptation mechanism achieve better performance than xNES with fixed learning rates?
\end{itemize}

We first describe the experimental setups in Section~\ref{sec:exp_setup}.
In Section~\ref{sec:evo_path_fixed}, we investigate the behavior of the evolution path in xNES with a fixed learning rate (RQ1).
We then investigate the behavior of the evolution path and the learning rate in xNES with the \emph{adaptive} learning rate mechanism (RQ2) in Section~\ref{sec:evo_path_adaptive}.
Finally, we compare the performance of xNES with the proposed adaptive learning rate mechanism and that with fixed learning rates (RQ3) in Section~\ref{sec:fixed_vs_adaptive}.
The code for running the proposed method is available at \href{https://github.com/nomuramasahir0/xnes-adaptive-lr}{\textbf{https://github.com/nomuramasahir0/xnes-adaptive-lr}}.

\subsection{Experimental Setups}
\label{sec:exp_setup}
Table~\ref{tab:benchmark} shows the definition of benchmark problems used in the experiment.
We employ two unimodal functions (Sphere and Ellipsoid) and two multimodal functions (Rastrigin and Bohachevsky).
While the Rastrigin function has strong multimodality, the Bohachevsky function has relatively weak multimodality.
In this experiment, we set the dimension to $d = 10$.
The initial parameters are set to $m^{0} = [3, \cdots, 3], \sigma^{(0)} = 2.0, B^{(0)} = I$ in the Sphere, Ellipsoid, and Rastrigin functions, and $m^{0} = [8, \cdots, 8], \sigma^{(0)} = 7.0, B^{(0)} = I$ in the Bohachevsky function.

The hyperparameters for the proposed learning rate adaptation mechanism are:
$\eta_{\sigma}^{\rm max} = \eta_{B}^{\rm max} = 1$, $\eta_{\sigma}^{\rm min} = \eta_{B}^{\rm min} = \frac{3}{5} \cdot \frac{(3 + \log(d))}{d\sqrt{d}}$, as described in Section~\ref{sec:update_lr}.
And we set $\alpha_{\sigma} = \alpha_{B} = 1.3$ and $\beta = \beta_{\sigma} = \beta_{B} = 0.2$ based on our preliminary experiments.
$\eta_{\sigma}^{\rm def}$ and $\eta_{B}^{\rm def}$ are set to their default values, i.e., 
$\eta_{\sigma}^{\rm def} = \eta_{B}^{\rm def} = \frac{3}{5} \cdot \frac{(3 + \log(d))}{d\sqrt{d}}$.

\begin{table*}[t]
  \centering
  \caption{Definition of benchmark problems.}
  \label{tab:benchmark}
  \begin{tabular}{l}
    \bottomrule
    Definition \\
    \hline \hline
    $f_{\rm Sphere}(x) = \sum_{i=1}^{d} x_i^2$  \\
    $f_{\rm Ellipsoid}(x) = \sum_{i=1}^{d} (1000^{\frac{i-1}{d-1}}x_i)^2$ \\
     $f_{\rm Rastrigin}(x) = 10d + \sum_{i=1}^d (x_i^2 - 10 \cos(2\pi x_i))$ \\
     $f_{\rm Bohachevsky}(x) = \sum_{i=1}^{d-1} ( x_i^2 + 2x_{i+1}^2 - 0.3 \cos(3\pi x_i) - 0.4\cos(4\pi x_{i+1}) + 0.7 )$ \\
    \bottomrule
  \end{tabular}
\end{table*}

\subsection{Evolution Path with Fixed Learning Rate}
\label{sec:evo_path_fixed}
\begin{figure}[tb]
  \centering
  \includegraphics[width=90mm]{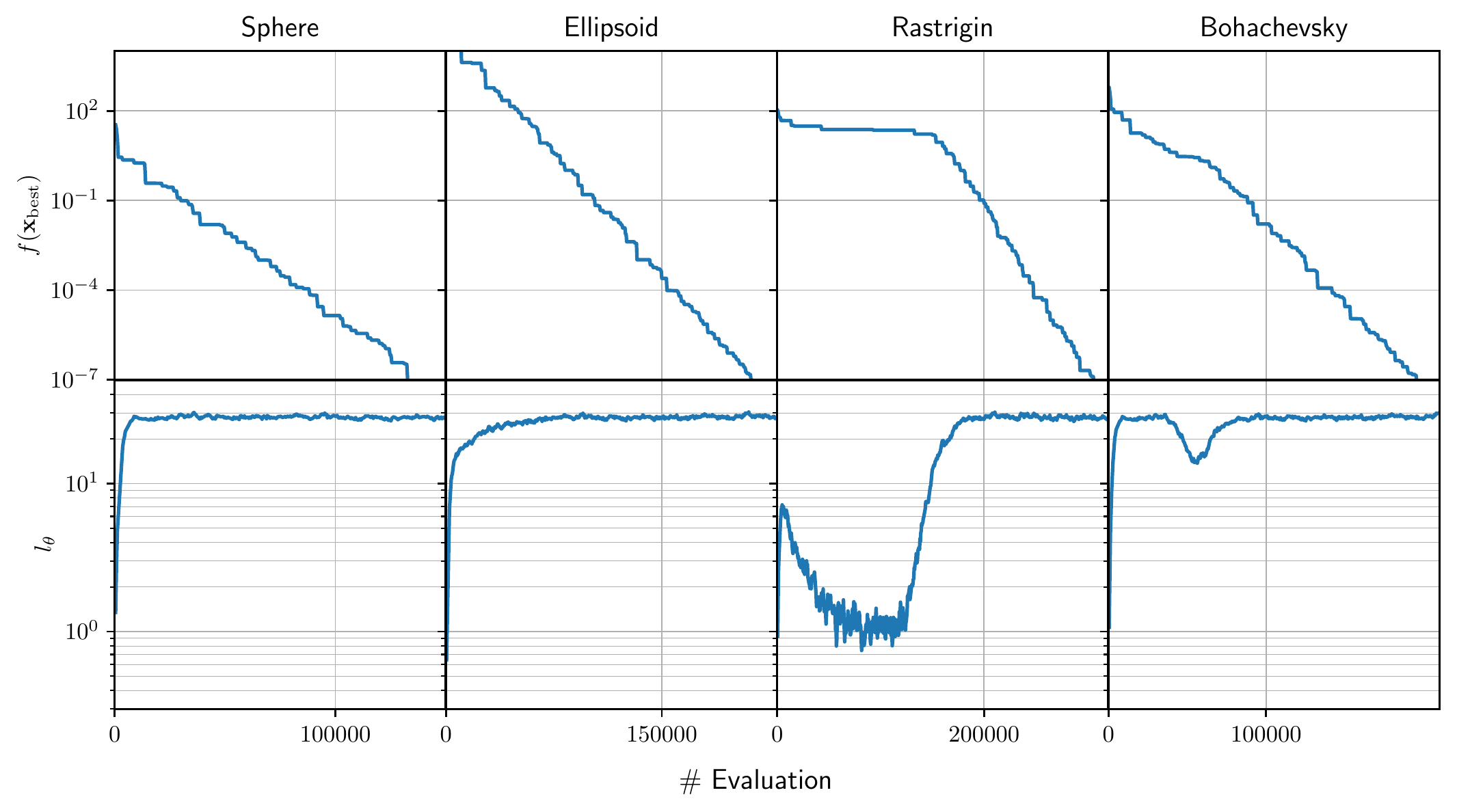}
  \caption{Typical behavior of xNES with a fixed learning rate on the $10$-dimensional benchmark problems. The horizontal axis represents the number of evaluations. The vertical axes represent the best evaluation value $f(x_{\rm best})$ and the length of the evolution path $l_{\theta}$, respectively.}
  \label{fig:fbest_kl_fixed_func}
\end{figure}

Figure~\ref{fig:fbest_kl_fixed_func} shows a typical behavior of the best evaluation value $f(x_{\rm best})$ and the length of the evolution path $l_{\theta}$ of xNES with a fixed learning rate on the benchmark problems.
We use the default learning rate and set the population size $\lambda = 400$ to obtain a reliable estimation of the evolution path.

In the result of the Sphere and Ellipsoid functions where $f(x_{\rm best})$ is improved quickly, the length of the evolution path $l(\theta)$ becomes long $(> 1)$.
We believe this is because the estimation accuracy of the natural gradient should be high in relatively easy objective functions (e.g., unimodal functions).

On the other hand, in the multimodal functions where $f(x_{\rm best})$ may not be improved easily, different behavior from that of the unimodal functions appears.
In the Bohachevsky function, which is a relatively weakly multimodal function, we can observe that the length of the evolution path slightly decreases once.
In the Rastrigin function, which has strong multimodality, such a decreasing behavior is prominent in the beginning of the optimization.
In fact, the length of the evolution path takes a value close to 1, which is the expected amount of change in KL divergence in a random function, in the period where the number of evaluations is between about $0.5 \times 10^5$ and about $1.2 \times 10^5$.

\subsection{Behavior of Learning Rate Adaptation}
\label{sec:evo_path_adaptive}
\begin{figure}[tb]
  \centering
  \includegraphics[width=90mm]{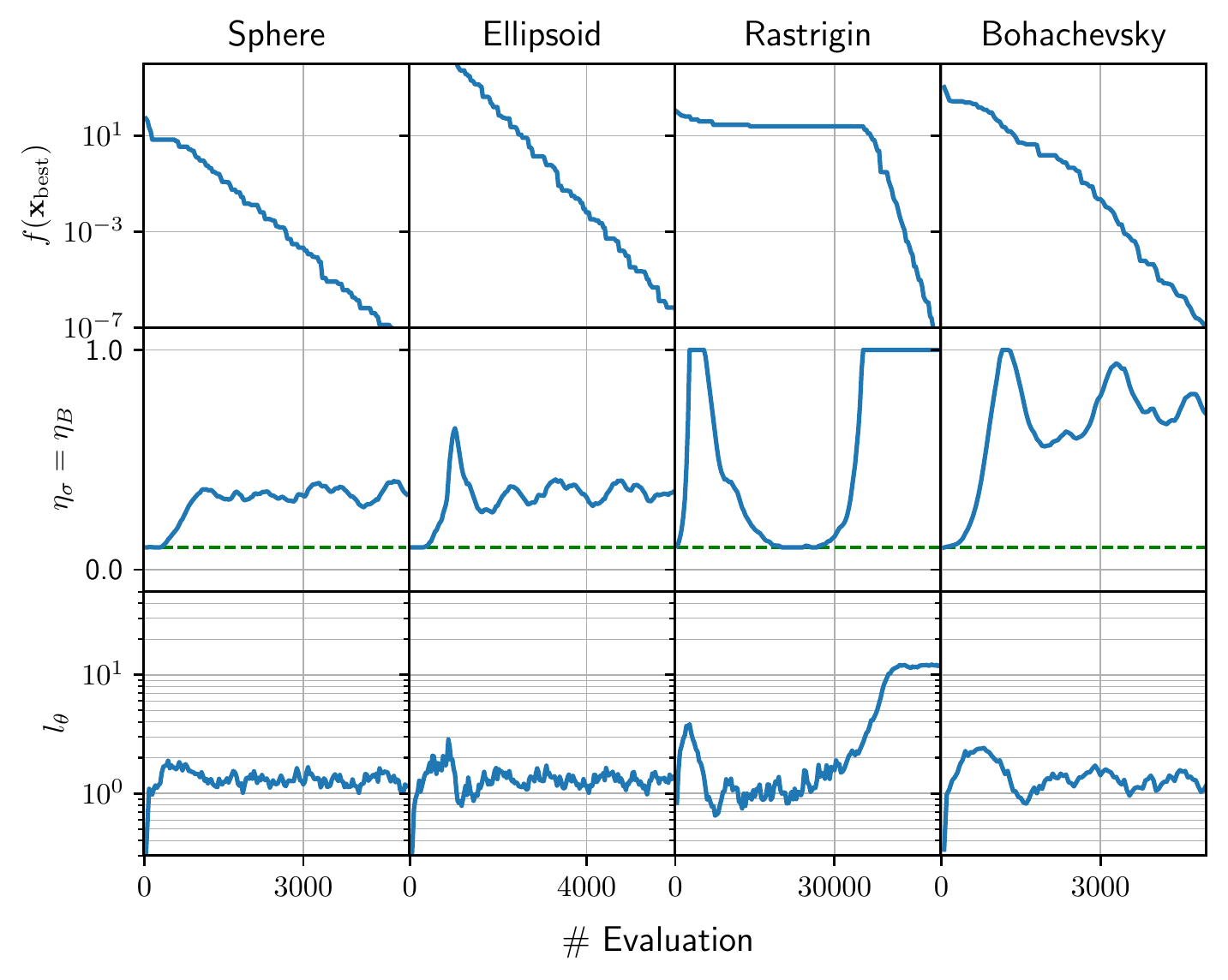}
  \caption{Typical behavior in xNES with the proposed learning rate adaptation mechanism on the $10$-dimensional benchmark problems. The green dotted line in the learning rate graphs indicates the default value. The horizontal axis represents the number of evaluations. The vertical axes represent the best evaluation value $f(x_{\rm best})$, the learning rates $\eta_{\sigma}$ and $\eta_B$, and the length of the evolution path $l_{\theta}$, respectively.}
  \label{fig:fbest_eta_kl_adaptive_func}
\end{figure}

A typical behavior of xNES with the proposed learning rate adaptation mechanism is depicted in Figure~\ref{fig:fbest_eta_kl_adaptive_func}.
In addition to the learning rates $\eta_{\sigma}$ and $\eta_B$, the corresponding objective function value $f(x_{\rm best})$ and the length of the evolution path $l_{\theta}$ are also shown.
We employ $\lambda = 30$ for the Sphere and Ellipsoid functions, $\lambda = 300$ for the Rastrigin function, $\lambda = 50$ for the Bohachevsky function, respectively.
It is observed in each function that the learning rates also increase when the length of the evolution path increases.

\begin{figure}[tb]
  \centering
  \includegraphics[width=90mm]{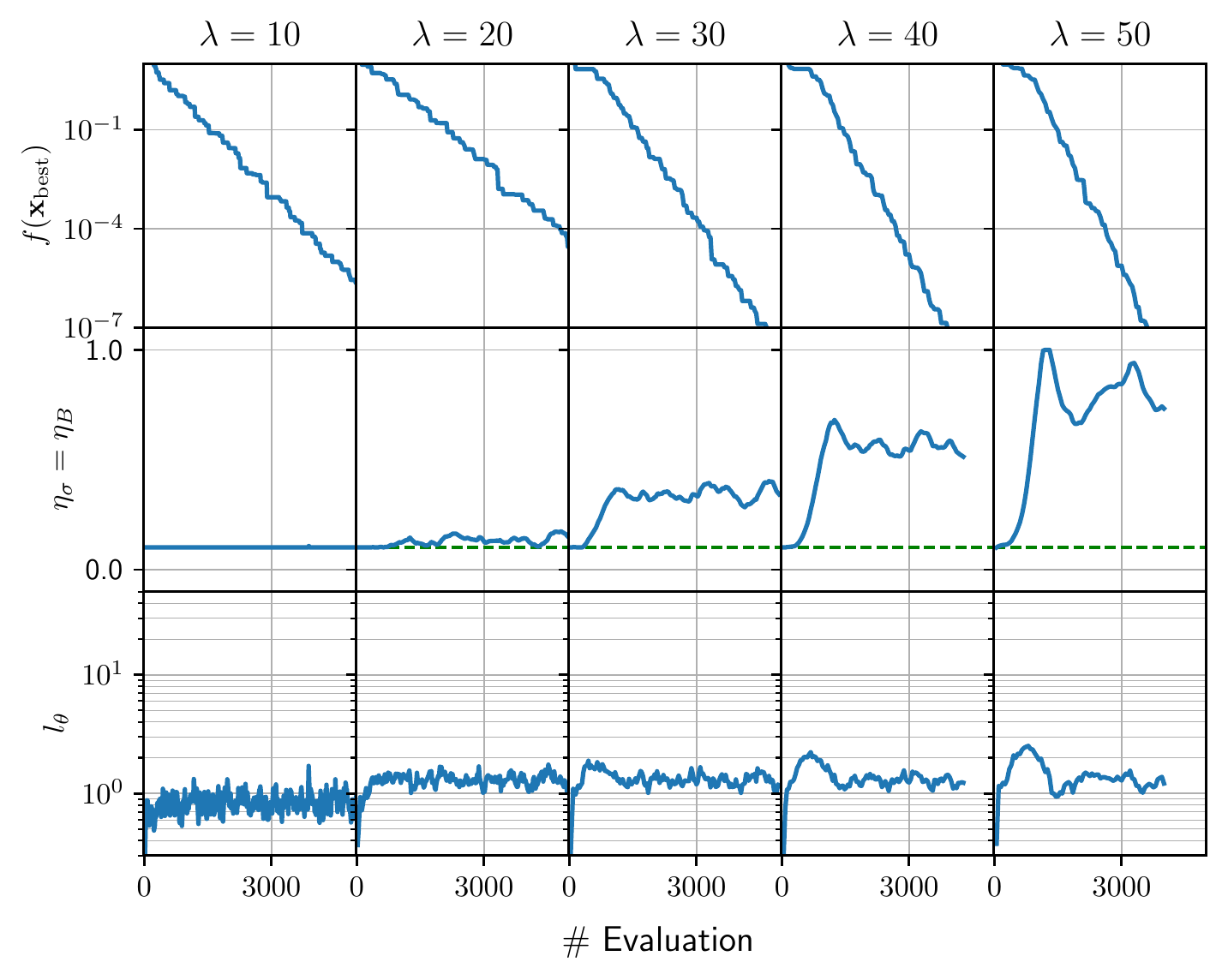}
  \caption{Typical behavior in xNES with the proposed learning rate adaptation mechanism on the $10$-dimensional Sphere function. The experiment is performed with the population size $\lambda = 10, 20, 30, 40$, and $50$. The green dotted line in the learning rate graphs indicates the default value.}
  \label{fig:sphere_fbest_eta_kl_adaptive_lambda}
\end{figure}

To investigate the effect of the setting of the population size, we conduct an experiment with $\lambda = 10, 20, 30, 40$, and $50$ on the $10$-dimensional Sphere function.
Figure~\ref{fig:sphere_fbest_eta_kl_adaptive_lambda} shows the result of the experiment.
In $\lambda = 10$, the length of the evolution path $l_{\theta}$ does not increase and the learning rates, $\eta_{\sigma}$ and $\eta_B$, are then not changed at all.
We think that this is because the accuracy of the parameter update is low under the setting of a small population size.
We observe that, as $\lambda$ is increased, the length of the evolution path increases, and, as a result, the learning rate also increases.
The result suggests that the proposed mechanism can adapt the learning rate appropriately, measuring the estimation accuracy of natural gradient.
This dynamic learning rate adaptation depending on the population size is an advantage over DX-NES~\cite{fukushima2011proposal}, which statically injects the population size into the setting of the learning rate.

\subsection{Fixed Learning Rate vs. Adaptative Learning Rate}
\label{sec:fixed_vs_adaptive}
To check the effectiveness of the proposed mechanism, we compare the performance of xNES with the proposed learning rate adaptation mechanism and that of xNES with fixed learning rates (= the default value $\times 1, 2, 4, 6, 8$, and $10$).
The performance metrics is the average number of evaluations until $f(x_{\rm best})$ reaches a target function value over successful trials divided by the success rate~\cite{auger2005restart}.
The target function value is set to $10^{-8}$.
A trial is successful if the target function value is found.
We set the maximum number of evaluations to $5 \times 10^5$.
For the Sphere function and the Ellipsoid function, we employ the population size $\lambda = 10, 20, 30, 40$, and $50$.
Note that the recommended value of the population size presented in~\cite{glasmachers2010exponential} is included, i.e., $4 + \lfloor 3 \ln(10) \rfloor = 10$.
For the Rastrigin function, we employ the population size $\lambda = 200, 250, 300, 350$, and $400$.
For the Bohachevsky function, we employ the population size $\lambda = 30, 40, 50, 60$, and $70$.
We perform $50$ trials to calculate the performance metrics for the Sphere and Ellipsoid functions.
We perform $200$ trials to calculate it for the Rastrigin and Bohachevsky functions.

\begin{figure}[tp]
  \centering
  \includegraphics[width=120mm]{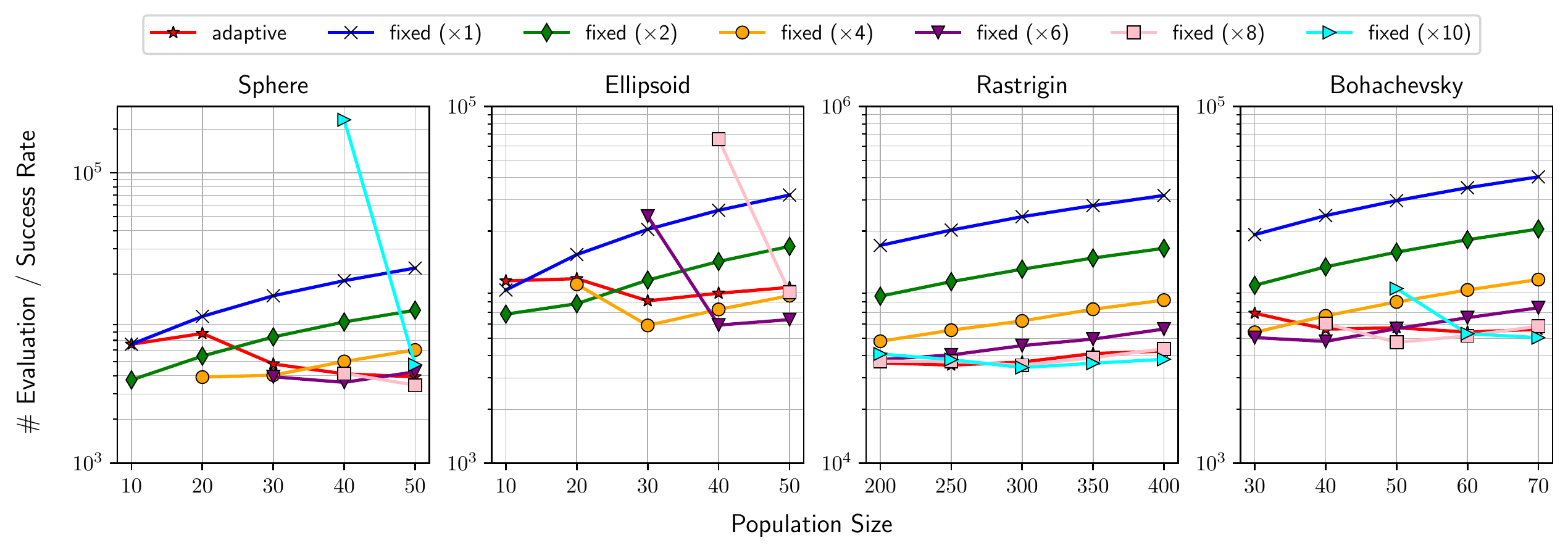}
  \caption{Performance comparison of xNES with the proposed learning rate adaptation mechanism (red) and xNES with the fixed learning rates (blue, green, yellow, purple, pink, and cyan) on $10$-dimensional benchmark problems. The horizontal axis represents the population size. The vertical axis represents the average number of evaluations divided by the success rate, which is the smaller, the better it is. Note that, if no successful trials exist at a population size, nothing is plotted at the population size.}
  \label{fig:main_compare}
\end{figure}

Figure~\ref{fig:main_compare} shows the result of the experiment.
We first compare the proposed mechanism (red) and xNES with the default learning rate (blue).
In the Sphere and Ellipsoid functions, when $\lambda = 10$, the performance is almost the same, which is consistent in Section~\ref{sec:evo_path_adaptive}.
As $\lambda$ increases, the proposed mechanism shows better performance than xNES with the default learning rate.
This is because that the estimation accuracy of the natural gradient should become high when $\lambda$ is large, which increases the learning rate and leads to acceleration of the search.
In the Rastrigin and Bohachevsky functions, the proposed mechanism outperforms xNES with the default learning rate due to the adaptive learning rate.

Next, we compare the proposed mechanism (red) and xNES with other fixed learning rates.
In all the benchmark problems, when $\lambda$ is large, the performance of the proposed mechanism is close to that of xNES with the fixed learning rate of the default value times $8$ (pink).
However, xNES with the fixed learning rate of the default value times $8$ fails to find the optimum in the Sphere and Ellipsoid functions with small population sizes ($\lambda = 10, 20$, and $30$) because the learning rate is too high.
On the other hand, the proposed mechanism does not increase the learning rate so much by measuring the estimation accuracy of the natural gradient when the population size is small, which enables stable search.

\begin{figure}[tp]
  \centering
  \includegraphics[width=100mm]{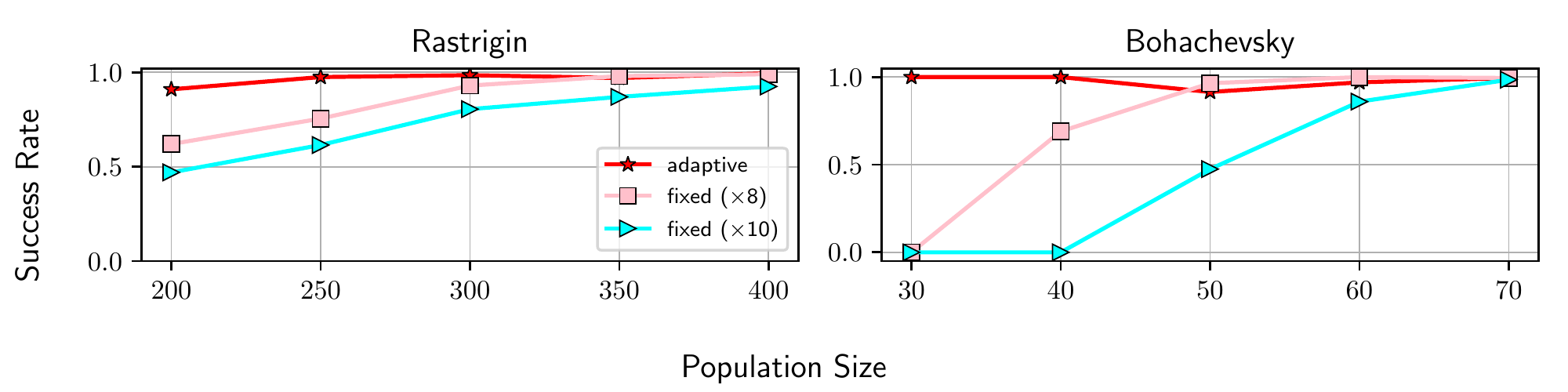}
  \caption{Success rate of xNES with the proposed learning rate adaptation method (red), xNES with the fixed learning rate of the default value times $8$ (pink), and xNES with the fixed learning rate of the default value times $10$ (cyan) in the multimodal functions.}
  \label{fig:succ_rate_multimodal}
\end{figure}

From the result in the multimodal functions, we can observe that the proposed mechanism is competitive with xNES with high learning rates when the population size is large.
In particular, in the Rastrigin function, the proposed mechanism and xNES with the fixed learning rate of the default value times $8$ and $10$ achieve almost the same performance in terms of the average number of evaluations of successful trials divided by the success rate.
This means that the number of evaluations required to find the optimum is about the same if an appropriate restart is performed when the optimum is failed to find.
Figure~\ref{fig:succ_rate_multimodal} shows the success rate of the proposed mechanism (red), xNES with the fixed learning rate of the default value times $8$ (pink), and xNES with the fixed learning rate of the default value times $10$ (cyan).
While these methods are competitive when the population size is large, xNES with the fixed learning rates are more likely to fail when the population size is small.
This result suggests that the proposed mechanism is more robust than xNES with a fixed learning rate.
A higher success rate in the proposed mechanism is also practically beneficial, as it is often difficult to implement an appropriate restart strategy.


\section{Conclusion}
\label{sec:con}

In this paper, we proposed a novel learning rate adaptation mechanism for NES.
The proposed mechanism adapts the learning rate based on the estimation accuracy of the natural gradient, which is inspired by the population size adaptation mechanism of the CMA-ES~\cite{nishida2016population,nishida2018psa}.
We introduced the evolution path in the parameter space of the covariance matrix.
Based on the length of the evolution path, we update the learning rates related to the covariance matrix.
The numerical experiments using unimodal and multimodal benchmark functions demonstrated that the proposed mechanism can appropriately adapt the learning rates depending on the estimation accuracy of the natural gradient.
Additionaly, xNES with the proposed mechanism achieved comparable performance as xNES with an appropriate fixed learning rate which cannot be obtained \emph{without} prior parameter surveys.

This study focused on the proposal of the principled learning rate adaptation, and did not conduct exhaustive experiments.
Verifying the performance of the proposed mechanism with a wider range of experimental setting is thus an important future direction.


\section*{Acknowledgement}
The authors thank anonymous reviewers for their helpful comments.
This work was partially supported by JSPS KAKENHI Grant Number JP20K11986.

\bibliographystyle{splncs04}
\bibliography{ref}

\end{document}